\documentclass[sigconf,screen]{acmart}

\usepackage[utf8]{inputenc}
\usepackage[T1]{fontenc}
\usepackage{graphicx}
\usepackage{colortbl}
\usepackage{tabularx}
\usepackage{multirow}  
\usepackage{pifont}
\usepackage{enumitem}

\acmSubmissionID{1137}

\renewcommand\footnotetextcopyrightpermission[1]{}

\begin{document}
\settopmatter{printacmref=false} 

\title{Towards Autonomous UAV Visual Object Search in City Space: Benchmark and Agentic Methodology}

\author{Yatai Ji}
\affiliation{%
  \institution{State Key Lab of Digital-Intelligent Modeling and Simulation}
  \city{Changsha}
  \country{China}}

\author{Zhengqiu Zhu}
\affiliation{%
  \institution{State Key Lab of Digital-Intelligent Modeling and Simulation}
  \city{Changsha}
  \country{China}}
\email{zhuzhengqiu12@nudt.edu.cn}

\author{Yong Zhao}
\affiliation{%
  \institution{State Key Lab of Digital-Intelligent Modeling and Simulation}
  \city{Changsha}
  \country{China}}

\author{Beidan Liu}
\affiliation{%
  \institution{State Key Lab of Digital-Intelligent Modeling and Simulation}
  \city{Changsha}
  \country{China}}

\author{Chen Gao}
\affiliation{%
  \institution{Department of Electronic Engineering, Tsinghua University}
  \city{Beijing}
  \country{China}}

\author{Yihao Zhao}
\affiliation{%
  \institution{Department of Electronic Engineering, Tsinghua University}
  \city{Beijing}
  \country{China}}

\author{Sihang Qiu, Yue Hu}
\affiliation{%
  \institution{State Key Lab of Digital-Intelligent Modeling and Simulation}
  \city{Changsha}
  \country{China}}


\author{Quanjun Yin}
\affiliation{%
 \institution{State Key Lab of Digital-Intelligent Modeling and Simulation}
  \city{Changsha}
  \country{China}
  }

\author{Yong Li}
\affiliation{%
  \institution{Department of Electronic Engineering, Tsinghua University}
  \city{Beijing}
  \country{China}
}


\begin{abstract}
Aerial Visual Object Search (AVOS) tasks in urban environments require Unmanned Aerial Vehicles (UAVs) to autonomously search for and identify target objects using visual and textual cues without external guidance. Existing approaches struggle in complex urban environments due to redundant semantic processing, similar object distinction, and the exploration-exploitation dilemma. To bridge this gap and support the AVOS task, we introduce CityAVOS, the first benchmark dataset for autonomous search of common urban objects. This dataset comprises 2,420 tasks across six object categories with varying difficulty levels, enabling comprehensive evaluation of UAV agents' search capabilities. To solve the AVOS tasks, we also propose \textbf{PRPSearcher} (\textbf{P}erception-\textbf{R}easoning-\textbf{P}lanning \textbf{Searcher}), a novel agentic method powered by multi-modal large language models (MLLMs) that mimics human three-tier cognition. Specifically, PRPSearcher constructs three specialized maps: an object-centric dynamic semantic map enhancing spatial perception, a 3D cognitive map based on semantic attraction values for target reasoning, and a 3D uncertainty map for balanced exploration-exploitation search. Also, our approach incorporates a denoising mechanism to mitigate interference from similar objects and utilizes an Inspiration Promote Thought (IPT) prompting mechanism for adaptive action planning. 
Experimental results on CityAVOS demonstrate that PRPSearcher surpasses existing baselines in both success rate and search efficiency (on average: +37.69\% SR, +28.96\% SPL, -30.69\% MSS, and -46.40\% NE). While promising, the performance gap compared to humans highlights the need for better semantic reasoning and spatial exploration capabilities in AVOS tasks. This work establishes a foundation for future advances in embodied target search.
Dataset and source code are available at https://anonymous.4open.science/r/CityAVOS-3DF8.
\end{abstract}






\begin{teaserfigure}
  \centering
   \includegraphics[width=0.8\linewidth]{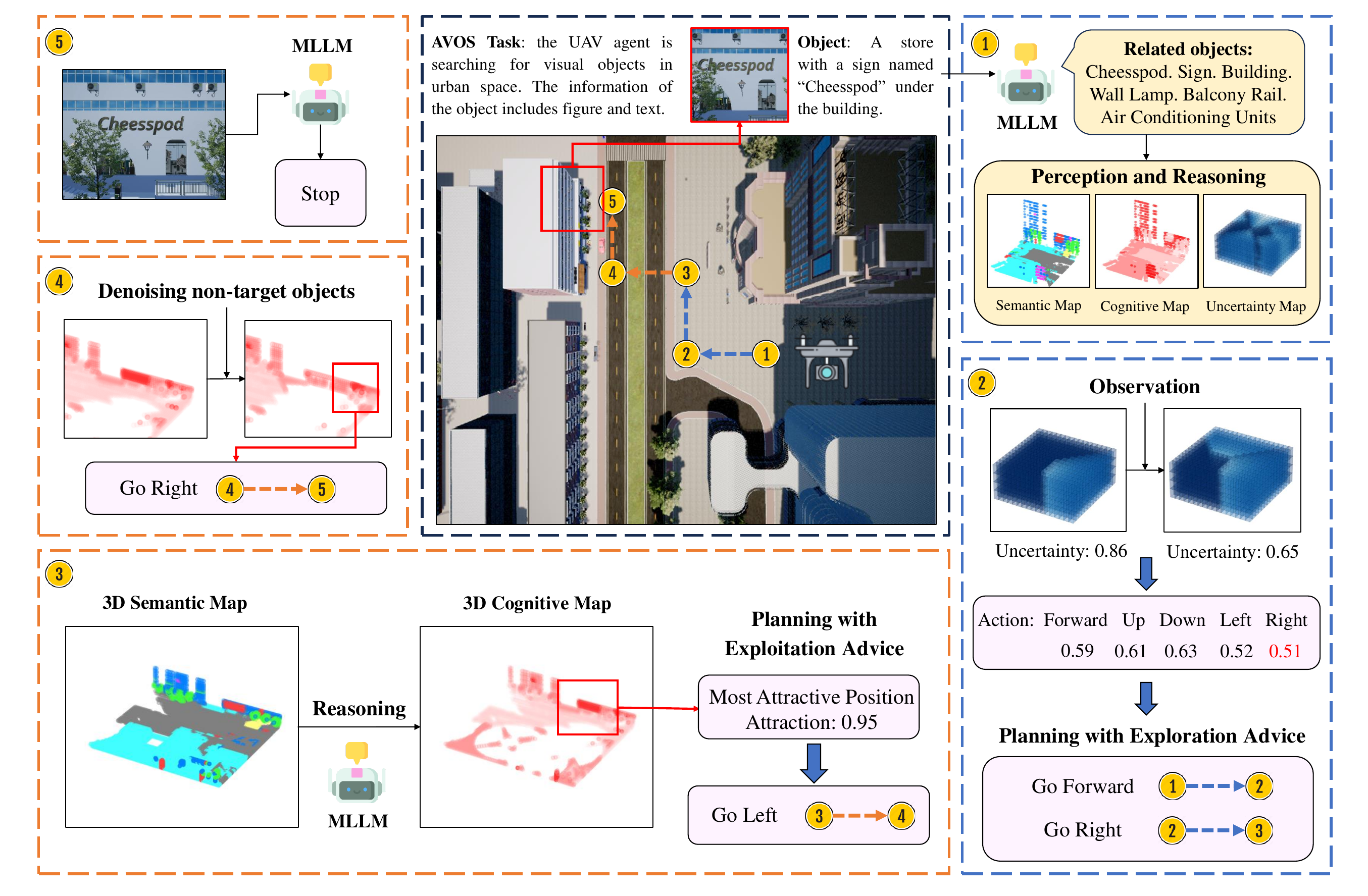}
   \caption{An illustration case of a UAV performing the AVOS task in an unfamiliar urban environment. In the search process, the UAV agent perceives the surrounding urban environments and reasons about the potential locations of the target object. In steps 1 and 2, the agent plans actions to explore the unknown space. In steps 3 and 4, the agent searches in the area with the highest attractions in the cognitive map. Finally, in step 5, the agent finds the target object and stops.}
   \label{fig:illustration}
\end{teaserfigure}

\maketitle  

\section{Introduction}

Unmanned Aerial Vehicles (UAVs) have found extensive applications in object search missions within city environments. Notable use cases encompass last-mile delivery in logistics systems ~\cite{she2021efficiency} and search operations in emergency response scenarios ~\cite{zhao2019uav}. Traditional solutions for UAV-based object search typically leverage metaheuristics or deep reinforcement learning methods to improve search efficiency through optimized flight path planning ~\cite{xing2022multi,hou2023uav}. However, the potential of dynamic visual observations is often overlooked. Recent advancements in embodied intelligence have enabled UAV-based agents driven by Multi-modal Large Language Models (MLLMs) to exhibit human-like proficiency in visual understanding, cognitive reasoning, and action decision-making ~\cite{liu2024aligning}. Consequently, the traditional object search task is transitioning towards Aerial Visual Object Search (AVOS) tasks, where UAVs are required to autonomously find visual objects in unfamiliar urban settings using provided cues (e.g., images, text descriptions, or both) without any navigational assistance or external instructions.

\begin{table*}
\centering
\caption{CityAVOS vs existing benchmarks. Datasets above the middle dividing line are the ground-based datasets, while those below are the aerial datasets. $N_{task}$: the number of tasks. $N_{traj}$: the number of total trajectories. Path Len: the average length of trajectories, measured in meters.}
\label{table:dataset}
\renewcommand\arraystretch{1.2}
\resizebox{0.65\linewidth}{!}{
\begin{tabular}{cccccccc}
    \hline
        ~ & Place & $N_{task}$ & $N_{traj}$ & Path Len. & Task Type & w/o Instruction  \\ \hline
        R2R~\cite{anderson2018vision} & Indoor (Ground) & 1020  & 7189 & 10.0 & Navigation &  \textcolor{red}{\ding{55}}   \\
        Reverie~\cite{qi2020reverie} & Indoor (Ground) & 4944 & 7000 & 10.0 & Navigation &  \textcolor{red}{\ding{55}}   \\ 
        ProcTHOR~\cite{deitke2022} & Indoor (Ground) & 10K  & - & - & Object Navigation &  \textcolor{green}{\ding{51}}    \\
        HM3DSem~\cite{yadav2023habitat} & Indoor (Ground) & 142646  & - & - & Object Navigation &  \textcolor{green}{\ding{51}} \\ \hline
        AerialVLN~\cite{liu2023aerialvln} & City (Aerial) & 8446 & 8446 & 661.8 & Navigation & \textcolor{red}{\ding{55}}   \\ 
        CityNav~\cite{lee2024citynav} & City (Aerial) & -  & 32637 & 545 & Navigation & \textcolor{red}{\ding{55}} \\ 
        EmbodiedCity~\cite{gao2024embodiedcity} & City (Aerial) & - & 99.7K & - & Navigation & \textcolor{red}{\ding{55}}  \\ 
        OpenUAV~\cite{wang2024towards} & City (Aerial) & - & 12149 & 255 & Navigation & \textcolor{red}{\ding{55}}  \\ 
        Openfly~\cite{gao2025openfly} & City (Aerial) & 3K & 100K & 99.1 & Navigation & \textcolor{red}{\ding{55}}\\ 
        CityAVOS (Ours) & City (Aerial) & 2420 & 2420 &  174.7& Object Search & \textcolor{green}{\ding{51}} \\ \hline
\end{tabular}}
\end{table*}

Currently, research on AVOS tasks within city spaces remains in its nascent stage. Tasks that bear resemblance to AVOS include vision-language navigation (VLN) ~\cite{lee2024citynav} and object goal navigation ~\cite{chen2023object, chaplot2020object} tasks, both of which leverage dynamic visual inputs to guide sequential action decisions. VLN tasks, which typically necessitate fine-grained navigation instructions to complete a specific trajectory, have been extended from indoor ~\cite{zhou2024navgpt} to outdoor scenarios, such as AerialVLN ~\cite{liu2023aerialvln}, OpenUAV ~\cite{wang2024towards}, and EmbodiedCity ~\cite{gao2024embodiedcity}. In contrast, AVOS tasks lack such fine-grained navigation instructions, instead relying on descriptions of target objects. Moreover, object goal navigation and AVOS tasks share a consistent task format, both aiming to locate specific objects in an unknown area. However, the majority of current research on object goal navigation predominantly focuses on indoor scenes \cite{wu2020active,wu2024voronav}.

This paper investigate the AVOS task in city spaces, which faces three unique challenges compared with previous studies:

1) \textbf{Complex and rich objects' semantics pose challenges to spatially-aware environmental representations}: 
Existing approaches primarily rely on point clouds or semantic grid maps for spatial awareness, but they often fall short in computational efficiency and mapping accuracy due to the redundant semantic information in complex urban environments. Therefore, a critical need exists for novel semantic mapping methods designed for urban contexts that are both computationally efficient and accurate.

2) \textbf{Similar objects' visual resemblance poses challenges to target reasoning and identification}: 
Urban scenes often feature multiple similar objects like shops, billboards, and cars, which are hard to distinguish remotely due to their visual resemblance. Accurate identification typically requires closer observation. Therefore, a key challenge lies in mitigating interference from these visually analogous yet incorrect targets during the target reasoning.

3) \textbf{Vast urban space and complex spatial structures pose challenges to action planning}: 
In large, complex urban settings, building, tree, and other occlusions can create visual blind spots in agent-constructed semantic maps. This leads to a difficult trade-off: searching only for semantic targets ignores unexplored areas, while exploring broadly is often inefficient.
Thus, balancing this exploration-exploitation dilemma in action planning is a challenge.

As an initial step, we develop a benchmark dataset, CityAVOS, to evaluate agents' performance on AVOS tasks. Tab. \ref{table:dataset} summarizes the differences between this dataset and other benchmark datasets. The CityAVOS dataset categorizes six target types and defines three levels of search difficulty. Task dataset involves searching for and identifying common urban targets by a UAV agent, described by both images and text descriptions, within complex scenes featuring intricate semantic information and spatial structures. Notably, UAV agents receive no guiding instructions, requiring them to perform a zero-shot autonomous search. Thus, the dataset evaluates their ability to autonomously search unfamiliar urban areas without other assistance.

To address AVOS tasks, we introduce PRPSearcher (\textbf{P}erception-\textbf{R}easoning-\textbf{P}lanning UAV \textbf{Searcher}), a novel agentic method powered by MLLMs, designed to mimic human three-tier cognition architecture for autonomous search of visual objects in urban spaces, as illustrated in Fig. \ref{fig:illustration}. 
\textbf{During the perception phase}, PRPSearcher extracts object-related semantics to construct the object-centric 3D dynamic semantic map. This map features object-centric semantic segmentation and a dynamic semantic label updating mechanism, which together enhance mapping efficiency and accuracy. Moreover, PRPSearcher constructs and updates a 3D uncertainty map to measure how much of the environment has been explored. 
\textbf{In the reasoning phase}, a 3D cognitive map is created based on "attraction values" (measures how strongly an object’s semantics attract a UAV agent) deducted by the MLLM. Moreover, we design a denoising mechanism to eliminate the influence of non-target objects. 
\textbf{In the planning phase}, we generate exploration and exploitation advice based on the cognitive map and the uncertainty map. Additionally, we introduce an Inspiration Promote Thought (IPT) prompting mechanism to help the agent strike a balance between exploration and exploitation during the decision-making process.
Results show that PRPSearcher achieves 53.50\% of SR and 40.57\% of SPL in CityAVOS tasks,  significantly surpassing the performance of baseline methods.

The contributions of this work are summarized as:
\begin{itemize}[leftmargin=*]
    \item To our knowledge, we are the first to introduce a benchmark dataset for the AVOS task in city space, namely CityAVOS.
    \item Inspired by human three-tier cognition, we propose an MLLM-based agentic method to address the AVOS task. This is achieved by constructing three types of maps — a semantic map, a cognitive map, and an uncertainty map — to enhance agents' spatial perception, target reasoning, and action planning capabilities.
    \item Experimental results demonstrate that our approach outperforms existing baselines in tackling the AVOS task. However, the gap with human performance highlights opportunities for future research to improve semantic reasoning and spatial exploration in embodied target search in city space.
\end{itemize}

\section{Related Work}

\subsection{Indoor Object Navigation}
The advent of simulators and datasets such as Matterport3D~\cite{chang2017matterport3d}, HM3D~\cite{ramakrishnan2021habitat} and Gibson~\cite{xia2018gibson} has driven significant progress in indoor navigation and search research~\cite{wijmans2019dd,ye2021auxiliary,ramakrishnan2022poni,chen2023object}. Early end-to-end methods~\cite{liu2020indoor,du2023object} directly mapped the observation to actions but incurred high computational costs. To mitigate this, Chaplot \textit{et al.}~\cite{chaplot2020object} proposed a graph-based modular method to integrate with learning-based approaches, reducing resource demands. 
Addressing zero-shot object navigation, Gadre \textit{et al.}~\cite{gadre2023cows} investigated the CLIP on Wheels (CoW) framework and benchmarks. Most recently, Large Language Models (LLMs) have been widely applied in the indoor object navigation methods~\cite{dorbala2023can,yu2023co,cai2024bridging}. For instance, L3MVN~\cite{yu2023l3mvn} used LLMs for commonsense reasoning to improve object search efficiency while ESC~\cite{zhou2023esc} transfers knowledge from pre-trained models for open-world object navigation.
VoroNav~\cite{wu2024voronav} presents a semantic exploration framework where an LLM leverages topological and semantic data to determine navigation waypoints. 

\textit{However, these studies primarily focus on indoor scenes, limiting their direct applicability to AVOS tasks in urban environments. However, their semantic mapping and cognitive reasoning approaches offer useful insights. These methods inspire us to develop outdoor exploration techniques that mimic human cognition, improving agents' spatial perception, target reasoning, and action planning capabilities.}

\vspace{-12pt}

\subsection{Urban Object Search}
Traditional urban object search methods~\cite{xing2022multi,hou2023uav} typically relied on optimization algorithms like meta-
heuristics~\cite{wu2023co} to generate search paths. Some other approaches incorporated Graph Neural Networks~\cite{zhang2025strucgcn} with Deep Reinforcement Learning~\cite{wu2019uav} to address this problem. However, these approaches often lack the capability to effectively process or incorporate visual object information. 
Recent advancements in embodied intelligence and Large Language Models (LLMs) have significantly propelled urban object search methodologies. For instance, Doschl \textit{et al.}~\cite{doschl2024say} proposed Say-REAPEx, an LLM-modulo online planning framework that prunes target-irrelevant actions from the planning process.  
To enhance LLM interpretability within urban contexts, NEUSIS~\cite{cai2024neusis} integrated neuro-symbolic methods to aid environmental reasoning. This progress is complemented by the rapid evolution of urban embodied environments and datasets, such as AerialVLN~\cite{liu2023aerialvln}, which provides a 3D simulator with near-realistic visuals for 25 city-scale scenarios, and the benchmark platform EmbodiedCity~\cite{gao2024embodiedcity} for embodied intelligence evaluation. Other outdoor embodied task platforms like OpenUAV~\cite{wang2024towards}, CityNav~\cite{lee2024citynav} and AeroVerse~\cite{yao2024aeroverse} also promoted the development of advanced urban object search methods. 

\textit{Nevertheless, there remains a notable absence of a dedicated AVOS benchmark tailored for urban environments, as well as a corresponding effective baseline model. Thus, this work contributes a comprehensive benchmark dataset for the AVOS task, and an effective MLLM-based agent baseline for autonomous visual search in urban environments.}

\begin{figure*}[!htp]
  \centering
   \includegraphics[width=\linewidth]{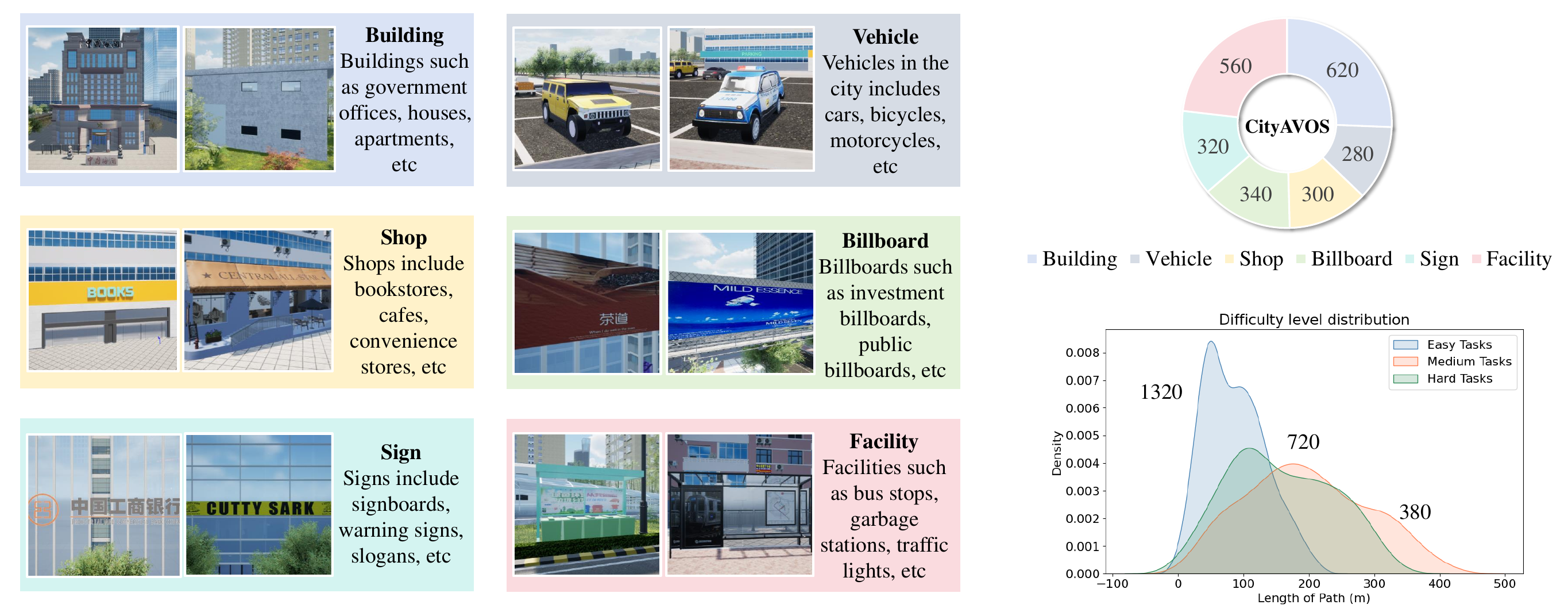}
   \caption{Examples of six object categories and dataset statistics of the CityAVOS.}
   \label{fig:datasetcate}
\end{figure*}
\section{CityAVOS Dataset}

In this section, we first define the AVOS task. Then, we introduce the simulated environment used to develop the CityAVOS dataset and outline the process of collecting and validating the dataset.

\subsection{Task Definition}
In an AVOS task $i$, a UAV agent is required to explore an unfamiliar urban environment and search for a visual object with task information $G_i$. At each step $t$, the agent perceives the RGB image $V_t$ and depth image $D_t$ in its current pose $P_t = [{pos_t,ori_t}]$. With observations $O_t = \{{V_t,D_t,P_t}\}$, the agent establishes an estimation of the visual object $E_t$. Then, a search policy $\pi(a_t|G_i,E_t)$ is employed to generate an action $a_t$. The agent determines whether to search and locate the target successfully based on observations. Finally, the search task ends when the agent executes the stop action.

\subsection{Dataset Collection}
We develop CityAVOS based on EmbodiedCity\cite{gao2024embodiedcity}, a platform built on Unreal Engine 5.3 that features high-fidelity simulations of urban streets, buildings, trees, vehicles, and pedestrians\cite{zhao2025cityeqa}. By integrating AirSim~\cite{shah2018airsim}, the platform provides a realistic environment for evaluating the performance of autonomous UAVs in urban settings. Using this environment, we define six distinct search scenarios (e.g., streets, neighborhoods, parks), with areas ranging from 5,600 to 82,800 square meters. To adapt these scenarios for the AVOS task, we embed specific recognizable objects within the scenes.

The dataset collection process consists of three main stages, involving both human operators and automated algorithms. The first stage is raw trajectory generation, which includes scene delimitation, target selection, and path collection. The second stage is task supplementation, involving the assignment of the agent’s initial pose and refinement of the corresponding task descriptions. Finally, the dataset undergoes validation and filtering to ensure quality and consistency. Further details are provided in Appendix~\ref{app.d1}.

\subsection{Dataset Statistics}

To further explore the proposed CityAVOS dataset, we demonstrate its characteristics from three aspects:
\begin{itemize}[topsep=0pt,parsep=0pt]
    \item \textbf{Construction of tasks:} Each task in CityAVOS is constructed as: $G  = (id, e, H, I, T, {P_{object}}, {P_0})$, where $id$ denotes the identity of an AVOS task, $e$ is the scene where the object exists, $H$ denotes the difficulty of the task, $I$ represents the visual information (image) of the object, $T$ represents the text information of the object, ${P_{object}}$ is the position of the object, and ${P_0}$ is the initial pose (including the 3D position and orientation) of the UAV agent.
    
    \item \textbf{Categories of objects:} The CityAVOS dataset contains 2,420 AVOS tasks and their corresponding trajectories, which consist of objects in the following six categories: \textit{building}, \textit{vehicle}, \textit{shop}, \textit{billboard}, \textit{sign}, and \textit{facility}. The distribution of these categories of tasks is illustrated in the top right corner of Fig. \ref{fig:datasetcate}.

    \item \textbf{Difficulty level of tasks:} The tasks in the dataset are categorized into three levels of difficulty: easy, medium, and hard. For easy tasks, the agent is required to locate a unique object within a small-scale scene. Medium tasks involve the agent searching for a unique object in a large-scale scene. Hard tasks require the agent to identify non-unique targets in a large-scale scene. The precise details regarding the difficulty classification are provided in Appendix \ref{app.d1}. The bottom right corner of Figure \ref{fig:datasetcate} illustrates the distribution of the corresponding difficulty levels.
\end{itemize}

\vspace{-8pt}
\section{The Agentic Method}

\begin{figure*}[t]
  \centering
   \includegraphics[width=0.8\linewidth]{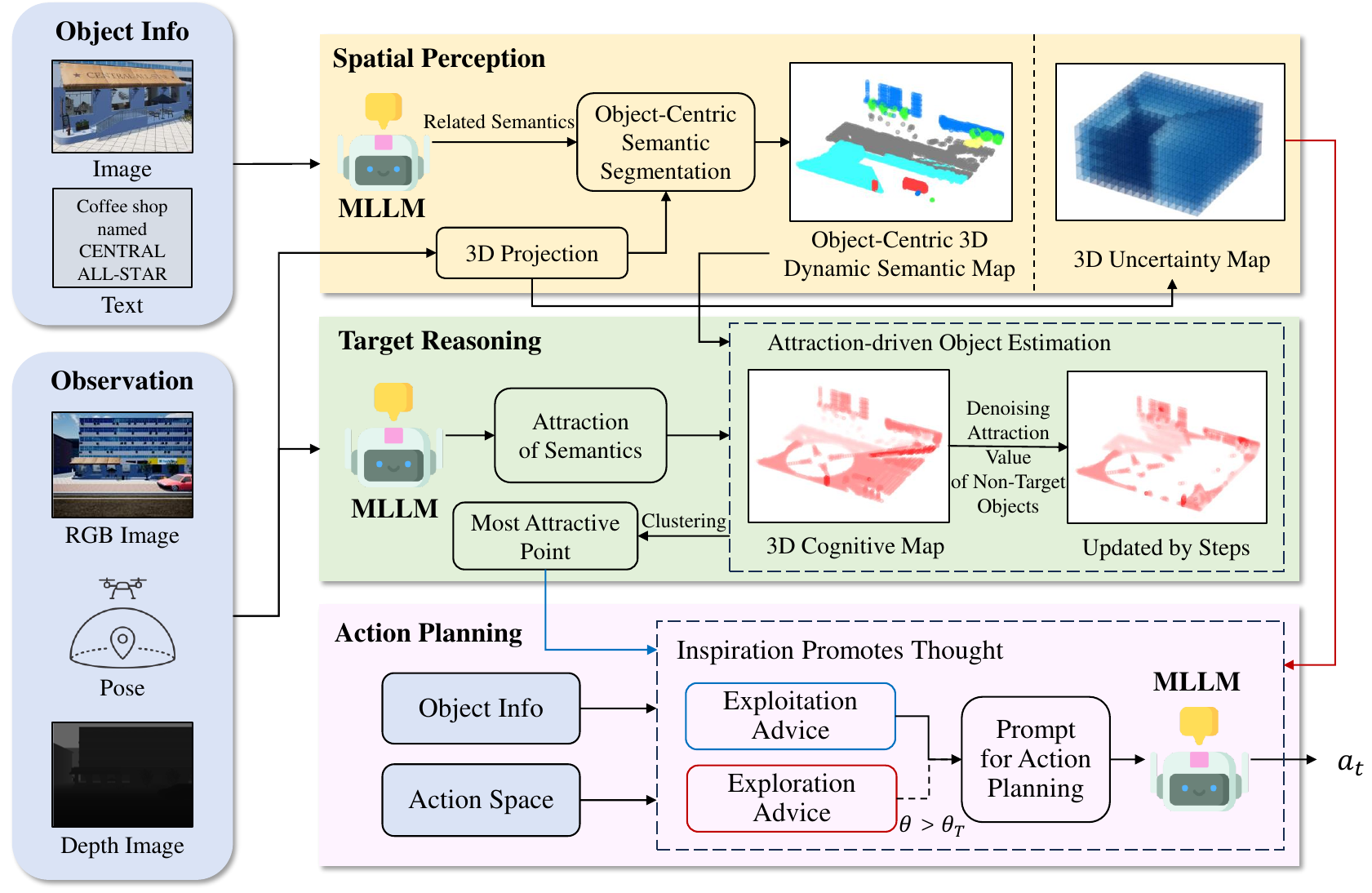}
   \caption{Overview of the agentic method--PRPSearcher.}
   \label{fig:overview}
\end{figure*}

\subsection{Overview}
An overview of the proposed PRPSearcher for the AVOS task is illustrated in Fig. \ref{fig:overview}, comprising three main phases: spatial perception, target reasoning, and action planning. 
\textbf{(1) In the perception phase}, the UAV agent creates an object-centric 3D dynamic semantic map of its surroundings by employing an MLLM to reason about target-related objects and extract corresponding semantics. This achieves object-centric semantic segmentation and reduces the computational costs for semantic mapping. Moreover, we adopt a dynamic-updating mechanism to improve mapping accuracy within the semantic grid. To quantify the extent of the environment explored in the current step, PRPSearcher also updates a 3D uncertainty map based on the UAV's visible area.
\textbf{(2) In the reasoning phase}, the UAV agent uses a 3D cognitive map to estimate the target's position. The map created by an MLLM is centered around the concept of "attraction." Attraction measures how strongly an object's semantics attract the UAV agent, based on that object's utility for finding the target. By clustering high-attraction grids within this map, the agent estimates the target's probable locations to guide its search plan. To ensure accuracy, a denoising mechanism mitigates the influence of objects unrelated to the target. 
Finally, \textbf{(3) in the planning phase}, we introduce the Inspiration Promotes Thought prompting mechanism for the UAV agent's action planning. This mechanism inputs target location estimates into the prompt as "exploitation advice", guiding the agent's search and target identification.  This is balanced by selectively adding "exploration advice" from a 3D uncertainty map, serving as "Inspiration" to encourage exploring unknown areas alongside exploiting known ones.


\vspace{-10pt}
\subsection{Object-Centric 3D Dynamic Semantic Map Construction Based on Spatial Perception}
To represent semantic distribution in urban environments, we construct an object-centric 3D dynamic semantic map (3D-grid form) based on visual observations and the UAV pose. 

\textbf{Object-Centric Semantic Segmentation.} For each task $i$, we employ an MLLM to reason about target-related objects based on task information (including image ${I_i}$ and text description ${T_i}$) and obtain relevant semantic elements:

\begin{equation}
E^i_s = \text{MLLM}({{Prompt}_{rel}}, {I_i}, {T_i}), 
  \label{eq:important1}
\end{equation}
where ${Prompt}_{rel}$ is the prompt input for an MLLM to generate the $E_s$. Details of the prompt can be found in Appendix \ref{app.approach}. 
These elements are integrated into the prompt for segmentation, serving the purpose of eliminating semantics unrelated to the target object. The semantic segmentation process is defined as:

\begin{equation}
S_s = Segment({E^i_s},V),
  \label{eq:important2}
\end{equation}
where $V$ is the RGB image from observation, $S_s$ denotes the results of semantic segmentation, including masks, boxes, and labels of each semantic element. $Segment()$ represents the semantic segmentation process by Ground-SAM model~\cite{bousselham2024grounding}.

\textbf{3D Dynamic Semantic Map.} Assuming the camera intrinsic matrix is $K \in \mathbb{R}^{3 \times 3}$ and the extrinsic matrix is $[R|r] \in \mathbb{R}^{3 \times 4}$, where $R$ is the rotation matrix and $r$ is the translation vector. For each pixel $ (u, v) $ in the depth image $D$, its world coordinates $(X, Y, Z)$ can be calculated using the following formula:

\begin{equation}
\begin{bmatrix}  
X \\
Y \\
Z  
\end{bmatrix}  
= R^{-1} \left( K^{-1} \begin{bmatrix}  
u \cdot D(u, v) \\
v \cdot D(u, v) \\
D(u, v)  
\end{bmatrix} - r \right) .
\end{equation}

Divide the world space into regular grids, each with a size of $ \Delta x \times \Delta y \times \Delta z $. For each pixel $ (u, v) $, its world coordinates $ (X, Y, Z) $ correspond to the grid indices $ (i, j, k) $: 

\begin{equation}
i = \left\lfloor \frac{X - x_{\text{min}}}{\Delta x} \right\rfloor, \quad j = \left\lfloor \frac{Y - y_{\text{min}}}{\Delta y} \right\rfloor, \quad k = \left\lfloor \frac{Z - z_{\text{min}}}{\Delta z} \right\rfloor,
\end{equation}
where $ [x_{\text{min}}, x_{\text{max}}] \times [y_{\text{min}}, y_{\text{max}}] \times [z_{\text{min}}, z_{\text{max}}] $ is the boundary of the scene space.  

For each grid $ (i, j, k) $, we count all the semantic labels of the pixels it contains, and select the most frequently occurring semantic label as the semantic representation of that grid: 

\begin{equation}
S_{i,j,k} = \arg\max_{c \in E_s} \sum_{(u, v) \in \text{pixels in } (i,j,k)} \mathbb{I}(L(u, v) = c),
\end{equation} 
where $ \mathbb{I}(\cdot) $ is the indicator function, $ L(u, v) $ is the semantics of the pixel $ (u, v) $ stored in the results of semantic segmentation $S_s$, and \textit{c} is the semantic category.


\subsection{Attraction-Driven Target Estimation}
Based on the 3D dynamic semantic map, we represent the agent's estimation of the target's position by constructing a 3D cognitive map. Additionally, a denoising mechanism is applied to eliminate interference from non-target objects during the search process.

\textbf{3D Cognitive Map.} The 3D cognitive map $C$ is a 3D grid map that is equal in size to the semantic map $S$. We employ an MLLM to measure how strongly an object’s semantics
attract the UAV agent. For each semantic category $c$, the attraction value is computed as:

\begin{equation}
A(s) = \text{MLLM}({{Prompt}_{att}}, {I_i}, {T_i}).
  \label{eq:important3}
\end{equation}

By calculating the attraction values $A(S_{i,j,k})$ for each grid $(i,j,k)$ in the semantic map, we can assign these values to the corresponding grids in the cognitive map:

\begin{equation}  
C_{i,j,k} = A(S_{i,j,k}).
\end{equation}

\textbf{Denoising Mechanism.} A mirrored cognitive map $C'$ is created to keep track of whether each grid has been recognized by the UAV agent. The state of each grid in $C'$ is represented as follows: 

\begin{itemize}
    \item $C'(i,j,k) = 1$. The grid $(i,j,k)$ has not been recognized.
    \item $C'(i,j,k) = 0$. The grid $(i,j,k)$ has been recognized.
\end{itemize}

When the UAV agent performs an observation action, it leverages its current position and viewing angle to determine which grid cells in the cognitive map are visible. For each visible grid $(i,j,k)$, if it is within the distance defined by the step size of the agent, it is updated in the mirrored cognitive map $C'$ as recognized: $C'(i,j,k) = 0$. 

To enhance the quality of the cognitive map by filtering out noise from recognized areas, we apply a denoising process using the mirrored cognitive map, formulated as below:

\begin{equation}  
C_{i,j,k} = C_{i,j,k}  \cdot C'(i,j,k).
\end{equation}  

\subsection{E-E Balanced Action Planning}
To find the target with higher efficiency and success rate, we need to achieve an exploration-exploitation balance in action planning.

\textbf{3D Uncertainty Map.} The 3D uncertainty map is also a three-dimensional grid map, where each cell $ (i, j, k) $ is associated with an uncertainty value $ U_{i,j,k} \in [0, 1] $. At the start of the search, all cells have an uncertainty value of 1, indicating complete uncertainty. 

A UAV agent performs an observation at position $ \mathbf{p} = (X, Y, Z) $ and orientation $ \mathbf{o} = (o_x, o_y, o_z) $. Based on the current position and orientation, the set of visible grid cells $ \mathcal{V} $ is computed. For each visible cell $ (i, j, k) \in \mathcal{V} $, we attenuate its uncertainty $ U_{i,j,k} $ based on distance. The uncertainty of different faces of a cell is calculated independently. The attenuation function $f(d)$ is defined as:  

\begin{equation}   
f(d) = e^{-\alpha \cdot d},  
\label{eq:9}
\end{equation} 
where $  d = \sqrt{(X - x_i)^2 + (Y - y_j)^2 + (Z - z_k)^2}  $ is the Euclidean distance from the grid cell $ (i, j, k) $ to the agent's position $ \mathbf{p} $,  $ \alpha $ is the attenuation coefficient, controlling the rate at which uncertainty decreases with distance. Thus, the updated uncertainty is: 
\begin{equation}  
U_{i,j,k}^{\text{new}} = U_{i,j,k}^{\text{old}} \cdot f(d).  
\label{eq:10}
\end{equation} 

Each time the agent performs an observation, the above process is repeated, and the 3D uncertainty map is updated as follows:  

\begin{equation}  
U_{i,j,k}^{\text{new}} = \begin{cases}  
U_{i,j,k}^{\text{old}} \cdot f(d) & \text{if } (i, j, k) \in \mathcal{V} \\
U_{i,j,k}^{\text{old}} & \text{otherwise}  
\end{cases}.  
\end{equation} 

\textbf{Exploration Advice.} Given the vast urban space, a UAV agent needs to explore more unknown areas to acquire information related to the target. To model the exploration process with the 3D uncertainty map, we define a reward function that quantifies the reduction in uncertainty achieved by each potential action within the agent's action space. The reward for an action is the total uncertainty reduction across all grid cells in the 3D Uncertainty Map.

The reward $ Reward(a) $ for an action $ a $ is defined as follows. Let $ \mathcal{A} $ be the set of possible actions available to the agent. For each action $ a \in \mathcal{A} $, the agent predicts the new position $ \mathbf{p}_a $ and orientation $ \mathbf{o}_a $ after executing the action. Based on $ \mathbf{p}_a $ and $ \mathbf{o}_a $, the set of visible grid cells $ \mathcal{V}_a $ is computed. Then, $ Reward(a) $ is computed as:  

\begin{equation}  
Reward(a) = \sum_{(i,j,k) \in \mathcal{V}_a} \left( U_{i,j,k}^{\text{old}} - U_{i,j,k}^{\text{new}} \right) ,
\end{equation}  
where $ U_{i,j,k}^{\text{new}} $ is the updated uncertainty for grid cell $ (i,j,k) $ after executing action $ a $, computed by formula \ref{eq:10}.  

The action that maximizes the reward can be formulated as:

\begin{equation}  
a^*_{exploration} = \arg\max_{a \in \mathcal{A}} Reward(a),  
\end{equation}  
where $a^*_{exploration}$ is the exploration advice for the agent.

\textbf{Exploitation Advice.} The 3D cognitive map reflects the "attraction" of these semantic elements to the search object. Areas with the highest attraction values are the most likely locations for the target object. Let $ \mathcal{G} $ be the set of high-relevance grids, defined as:  
\begin{equation}   
\mathcal{G} = \{ (i, j, k) \mid C_{i,j,k} = max(C_{i,j,k}) \}.  
\end{equation}  

By using the DBSCAN clustering method~\cite{schubert2017dbscan}, several clusters $ \mathcal{C}_1, \mathcal{C}_2, \dots, \mathcal{C}_n $ can be identified as high-relevance regions. For the largest cluster $ \mathcal{C}_m $, the center point $ \mathbf{p}_m = (X_m, Y_m, Z_m) $ is calculated as the target point for the exploitation process. The action $a^*_{exploitation}  $ that navigates to the point $\mathbf{p}_m$ is the generated exploitation advice for the UAV agent.

\textbf{IPT-based E-E Balanced Planning.} 
In search tasks, exploration involves searching unfamiliar environments to gather new information, while exploitation relies on existing knowledge to estimate the target object’s location. Striking an optimal balance between these two modes is a critical challenge, as it is often difficult to determine whether the agent should act based on exploration or exploitation advice. When humans search for objects, they typically begin by considering the most likely locations of the target and then investigate those areas thoroughly. During the process, spontaneous thoughts such as "There's a place I haven't checked yet" often arise—this type of inspiration helps avoid overlooking potential locations. Such behavior reflects a natural balance between exploration and exploitation in human cognition. Motivated by this insight, we replicate this cognitive process by proposing the IPT prompting mechanism, which stimulates "inspirational" thinking in UAV agents to achieve a balanced exploration-exploitation (E\&E) strategy. An example of the prompt is provided in Appendix~\ref{app.approach}.

This mechanism integrates exploitation advice as long-term guidance into the agent's action planning prompt. This advice will continuously guide the agent in finding and identifying known objects. In contrast, exploration advice will be selectively incorporated into the prompt in the form of "Inspiration". There are several conditions in the search process where the agent should favor an exploration strategy: during the initial search phase or when the search becomes stuck in a local optimum. To facilitate this, we introduce a threshold $\theta$ to assess whether the benefits of exploration actions are significant enough. When the benefits exceed this threshold, exploration advice will be added to the planning prompt to remind the agent to shift its focus toward exploring unknown spaces.
\begin{equation}
{Prompt}_{plan} = {Advice}_{exploit} + I(Reward(a^*)>\theta) \cdot {Advice}_{explore}
\end{equation}
where $I()$ is the Boolean function, $I(Reward(a^*)>\theta)=1$ when $Reward(a^*)>\theta$ is true, otherwise $I(Reward(a^*)>\theta)=0$.

The numerical experiments related to parameter $\theta$ can be found in section \ref{section:Ablation Study}.

\section{Experiments}

\subsection{Experiment Setup}

\begin{table*}[t]  
    \centering  
    \caption{Performance comparisons with SOTA baselines on CityAVOS benchmark.}  
    \resizebox{\linewidth}{!}{
    \begin{tabular}{@{}lcccc|cccc|cccc|cccc@{}}  
        \toprule  
       \textbf{ Method} & \multicolumn{4}{c|}{Easy Tasks} & \multicolumn{4}{c|}{Medium Tasks} & \multicolumn{4}{c}{Hard Tasks} & \multicolumn{4}{c}{Total Tasks} \\ \cmidrule(lr){2-5} \cmidrule(lr){6-9} \cmidrule(lr){10-13} \cmidrule(lr){14-17}      
        & SR$\uparrow$ & MSS$\downarrow$& SPL$\uparrow$ & NE$\downarrow$& SR$\uparrow$ & MSS$\downarrow$& SPL$\uparrow$ & NE$\downarrow$& SR$\uparrow$ & MSS$\downarrow$& SPL$\uparrow$ & NE$\downarrow$&
        SR$\uparrow$ & MSS$\downarrow$& SPL$\uparrow$ & NE$\downarrow$\\ \midrule 
        \rowcolor{gray!20} 
        Human     &     \textbf{85.45}&     \textbf{17.40}&    \textbf{76.58}&  \textbf{20.74}&       \textbf{72.16}&       \textbf{17.76}&    \textbf{68.31}&  \textbf{56.50}&    \textbf{67.68}&    \textbf{15.94}&                     \textbf{56.71}& \textbf{31.43}&   \textbf{78.68}&   \textbf{17.26}&   \textbf{70.92}&  \textbf{32.90}\\ \hline
        RE    & 10.30& 49.35&     6.90&     89.00& 3.98& 62.07&     1.82&     198.23& 7.07& 97.75&   3.69&  153.96& 7.93&60.97&   4.90&    131.41\\
FBE     & 13.64& 39.47&    10.04&  97.48& 9.66& 58.85&    7.67&  194.71& 5.05& 60.93&                      3.81&  198.38& 11.07& 48.62&   8.33& 142.33\\ \
L3MVN     &  26.82&  34.51&  21.54&  87.89&  7.09&  60.02&  4.06&  190.34&  7.21&   59.68&  3.94&  180.84&   17.87&   46.05&   13.57&  132.90
\\
WMNav &       20.62&       38.54&       18.05&       75.06&       5.42&       69.86&       3.19&       164.69&       12.17&       77.35&                           8.72&   110.79&   14.82&   54.02&   12.20&     106.99 \\
STMR &       32.68&       34.25&       23.86&       66.07&       21.52&       55.68&       13.9&       138.66&       19.91&       60.19&                           11.96&   89.41&   27.35&   44.70&   19.03&     91.33 \\
\midrule
\rowcolor{blue!5}
PRPSearcher w/o exploitation& 16.36& 38.87&  13.25&  95.25& 3.41& 59.61&  2.11&  165.28& 3.03& 61.66&                     2.38& 101.06&10.41& 48.63&   8.23&  116.57\\
\rowcolor{blue!15}
PRPSearcher w/o exploration& 60.47& 30.22&  47.89&  50.19& 39.68& 45.86&  35.09&  129.47& 28.52& 46.08&                     16.68& 92.36&49.19& 37.37&   39.06&  80.16\\
\rowcolor{blue!30} 
PRPSearcher& \underline{66.32} & \underline{28.85}&  \underline{49.82}&  \underline{43.62}& \underline{42.89} & \underline{41.33} &  \underline{36.68} &  \underline{98.35} & 
\underline{29.62} & \underline{45.84}&      \underline{16.65}& \underline{76.13} & \underline{53.50} & \underline{35.26} &  \underline{40.57} &  \underline{64.86}  \\
 \bottomrule  
    \end{tabular}  }
    \label{tab:performancemetrics}  
\end{table*}

\textbf{Evaluation Metrics.} We adopt four standard metrics to measure the performance, i.e., Success Rate (SR), Success Rate Weighted by Inverse Path Length (SPL)~\cite{wu2024voronav}, Mean Search Steps (MSS)~\cite{zhao2022deep}, and Navigation Error (NE)~\cite{ramakrishnan2022poni, liu2023aerialvln}. The details of the four metrics can be found in Appendix \ref{app.metric}.
SR calculates the percentage of episodes in which the agent terminates within a predefined success threshold (20 meters) and successfully identifies the target. SPL measures navigation efficiency as the inverse ratio of the actual path length to the optimal path length, weighted by success rate. The path length is calculated as the cumulative distance between consecutive actions. MSS, often used in object search tasks, represents the average number of actions that the agent takes in each episode. NE measures the Euclidean distance between the final position of the agent and the ground truth target object.

\textbf{Implementation Details.} For PRPSearcher, the input image is resized to 640 × 480 for convenient processing, and some commonly used MLLMs (e.g., GPT-4o and Qwen-vl-max) are leveraged for visual analysis and reasoning during the spatial perception, target reasoning, and action planning phases. The dataset used for the experiment is CityAVOS, and the platform is the EmbodiedCity modified for AVOS. Due to API limitations, 605 tasks (25\%) are randomly selected from the CityAVOS dataset for extensive experiments. 

\textbf{Baselines.} Our baseline comparisons utilize object search studies from the last two years, encompassing both indoor and outdoor research. Furthermore, acknowledging the nascent nature of the AVOS task, we supplement these with foundational methods to ensure a comprehensive performance evaluation.

\begin{itemize}[leftmargin=*]
    \item \textbf{Random Exploration (RE):} The agent randomly selects one action to execute until the ‘stop’ action is chosen.
    \item \textbf{Frontier-Based Exploration (FBE):} A purely frontier exploration method that ignores semantic information~\cite{ramakrishnan2022poni}.
    \item \textbf{L3MVN:} L3MVN~\cite{yu2023l3mvn} records semantic information on the frontiers of a frontier map and leverages LLMs to determine which frontier to prioritize for object search.
    \item \textbf{WMNav:} WMNav~\cite{nie2025wmnav} constructs a curiosity value map to predict the likelihood of the target’s presence. Direction of the highest value is selected and sent to the navigation policy module.
    \item \textbf{STMR:} STMR~\cite{gao2024aerial} extracts instruction-related semantic masks of landmarks into a top-down map for action prediction.
    \item \textbf{Human Agent:} The actions of the UAV are determined by an individual human participant based on real-time observations obtained from the UAV. Five postgraduates with drone-operating expertise participated in the experiment, though all lacked familiarity with urban environments. Results reflect the average performance across participants.
\end{itemize}

To adapt the baseline indoor object search methods for urban outdoor settings, we have made some adjustments to these methods, including (but not limited to) input matching and converting 2D structures into 3D structures.
For outdoor research, Say-REAPEx~\cite{doschl2024say} and NEUSIS~\cite{cai2024neusis} represent the latest studies related to object search. However, as these methods are not currently open-sourced and key components are challenging to replicate, we have excluded them from the baselines in this study. Additionally, for benchmarks such as OpenUAV~\cite{wang2024towards} and OpenFly~\cite{gao2025openfly}, the methods they proposed are based on their own trained models, which are not applicable to the AVOS task. As a result, these methods have also been omitted from the baselines.

More details about the experimental implementation and results can be found in Appendix \ref{app.experiment}.

\subsection{Comparisons with SOTA Methods}

As shown in Tab. \ref{tab:performancemetrics}, our proposed approach significantly outperforms the baseline methods (on average: +37.69\% SR, +28.96\% SPL, -30.69\% MSS, and -46.40\% NE) in tasks of all difficulties, demonstrating the effectiveness of the designed mechanisms and the constructed maps. However, the gap with human performance indicates that the reasoning capabilities of existing MLLMs, along with other mechanisms designed in this work, are still insufficient to match those of human operators.
Some observations can be obtained: 

\begin{itemize}[leftmargin=*]
    \item \textbf{Basic Method.} The random exploration method and frontier-based exploration methods perform poorly on tasks of various difficulties. As both types of methods are blind space exploration approaches, their performance reflects that the AVOS tasks cannot be solved through basic space exploration patterns.

    \item \textbf{Indoor Method.} Although the success rate of indoor methods is not high, there is a significant improvement compared to the basic method. The L3MVN method has increased the success rate (SR) by 13.18\% on the simple difficulty task set compared to the basic method. The WMNav method achieves good performance on hard tasks through a curiosity mechanism. These results not only highlight the importance of understanding semantics for AVOS tasks but also reflect the limitations of indoor methods in city environments.

    \item \textbf{Outdoor Method.} The STMR method performs best in baselines except for the human agent. STMR facilitates the storage of outdoor semantic information by constructing a Top-down map in the air. Meanwhile, it enhances the ability of agent action planning based on the Chain of- Thought reasoning. Therefore, the SR in medium and hard tasks can reach 21.52\% and 19.91\% respectively. This result reflects the importance of the reasoning ability of agents in highly difficult tasks.

    \item \textbf{Human Agent.} Human agents performed best in all task classifications, thanks to humans' innate strong visual understanding and sequential action decision-making abilities. With the increase of task difficulty, the performance of human agents also decreased slightly, which indicates that there are certain challenges for human beings to successfully complete AVOS tasks. The proposed PRPSearcher achieves 68\% human-level performance on SR, which illustrates the advanced nature of the approach and also hints at the potential for further performance improvements on AVOS tasks.

\end{itemize}

Overall, the comparisons with baseline models reveal that \textit{excluding interference from redundant object information during semantic extraction and effectively distinguishing target-like objects in urban environments} are crucial for improving search efficiency. Additionally, achieving a higher success rate in AVOS tasks \textit{depends on striking an optimal balance between exploitation and exploration}.

\begin{figure*}[!htp]
  \centering
   \includegraphics[width=0.85\linewidth]{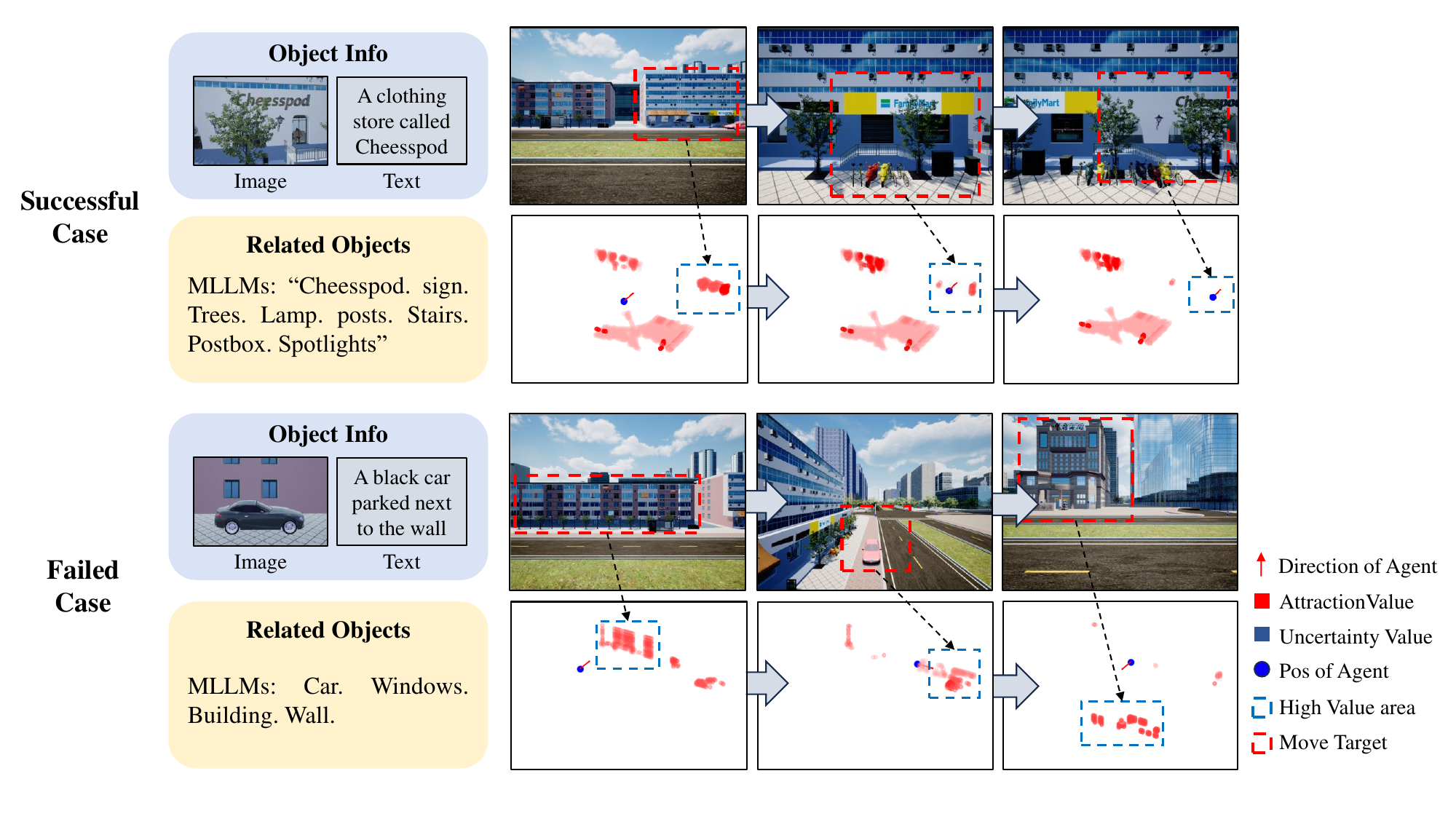}
   \caption{Two selected cases of PRPSearcher on two episodes. One is a successful case, demonstrating how and why our proposed approach effectively finds and identifies the target compared to baseline models. While another one is a failed case, highlighting the limitations of our approach in visual reasoning capability compared to human agents.}
   \label{fig:case}
\end{figure*}

\subsection{Ablation Study}
\label{section:Ablation Study}

\textbf{Effect of the object-centred 3D dynamic semantic map.} To manifest the contribution of the object-centered 3D dynamic semantic map proposed in this paper to the spatial perception of the agent, we conduct ablation experiments and design two other semantic segmentation prompts: free-prompt and human-designed. The former does not provide prompts to the semantic segmentation model, allowing the model to determine the segmentation targets on its own. The latter involves humans actively setting the prompts, without further adjustments for different tasks. The experimental results indicated in Tab. \ref{tab:ablationsemantic} show that the human-designed semantic segmentation prompts achieved the worst experimental results, and the performance of the free-prompt is slightly lower than our method. When designing the semantic segmentation prompts, we used the dataset’s classification labels for the search targets as inputs to the semantic segmentation model. This led to overly rich semantics in the semantic map, which somewhat interfered with the agent’s judgment. Similarly, when using the semantic segmentation model to perform segmentation autonomously, it also introduced a large amount of semantics from non-target objects, reducing both search efficiency and success rate.

\begin{table}[t!]  
    \centering  
    \caption{Ablation study of the object-centred 3D dynamic semantic map for PRPSearcher.}  
    \begin{tabular}{lcccc}
\hline
\textbf{Method} & \multicolumn{4}{c}{Total}     \\ \cline{2-5} 
                        & SR$\uparrow$ & MSS$\downarrow$& SPL$\uparrow$ & NE$\downarrow$\\ \hline
free-prompt             &  50.52&  37.89&  38.11&  84.03
\\
human-design             &  38.46&  41.41&  30.27& 105.68
\\
\rowcolor{blue!30} 
object-centric            & \textbf{53.50}& \textbf{35.20}&  \textbf{40.57}&  \textbf{64.86}
\\ \hline
\end{tabular}
    \label{tab:ablationsemantic}  
\end{table}

\textbf{Effect of the exploration and exploitation design.} The approach proposed in this paper achieves a balance in action planning through providing the agent with exploration advice and exploitation advice. Specifically, the exploration advice is derived from the 3D uncertainty map, while the exploitation advice comes from the 3D cognitive map. Therefore, this ablative experiment aims to validate the contributions of these two map designs. The experimental results are shown in Tab. \ref{tab:performancemetrics}. The \textbf{PRPSearcher w/o exploration} method still maintains a high performance, but the absence of suggestions for exploring unknown spaces results in a decline in both SR and SPL. Conversely, the \textbf{PRPSearcher w/o exploitation} method performs poorly, yet still outperforms the FBE method. This further demonstrates the importance of semantic understanding for the AVOS (Autonomous Visual Object Search) task. At the same time, the above experimental results confirm the effectiveness of the method proposed in this paper.

\textbf{Effect of the IPT prompting mechanism.}
The IPT prompt mechanism is designed to balance exploration and exploitation during the agent’s action planning. A key parameter in this mechanism, denoted as $\theta_T$, controls the frequency of exploration advice provided to the agent. We conducted numerical experiments to evaluate the impact of different $\theta_T$ values, and the results are summarized in Table \ref{tab:ablationIPT}. When $\theta_T = 0.5$ or $\theta_T = 1$, the number of exploration prompts received by the agent drops to zero ($N_{\theta} = 0$), leading to a decline in performance due to the lack of exploratory guidance. Conversely, when $\theta_T = 0$, the agent receives exploration advice at every decision step, which overwhelms its decision-making process and significantly reduces the success rate (SR). Through these experiments, we identified $\theta_T = 0.1$ as the optimal setting, effectively enabling the agent to strike a balance between exploration and exploitation during action planning.

\begin{table}[t!]  
    \centering  
    \caption{Ablation study of the IPT prompting mechanism for PRPSearcher.}  
\begin{tabular}{cccccc}
\hline
$\theta_T$                       & SR            & MSS            & SPL            & NE             & $N_{\theta}$     \\ \hline
1                           & 49.19         & 37.37          & 39.06          & 80.16          & 0     \\
0.5  & 49.38         & 38.44          & 38.83          & 78.62          & 0     \\
0.2  & 51.38         & 35.3           & 39.89          & 67.78          & 4.37  \\
\rowcolor{blue!30} 
0.1  & \textbf{53.5} & \textbf{35.26} & \textbf{40.57} & \textbf{64.86} & 8.62  \\
0.05 & 43.99         & 41.71          & 32.48          & 89.47          & 26.09 \\
0.02 & 38.2          & 44.59          & 30.05          & 97.58          & 44.59 \\
0                           & 38.37         & 44.82          & 29.9           & 95.11          & 44.82 \\ \hline
\end{tabular}
    \label{tab:ablationIPT}  
\end{table}

\textbf{Effect of the different MLLMs.} As PRPSearcher is an MLLM-based agentic methodology, we further evaluate the abilities of different MLLMs in AVOS taks as shown in Tab. \ref{tab:performancellm}. The experimental results show that the PRPSearcher exhibits good search performance under different MLLMs (Multimodal Language Models) loads. Among the three MLLMs, glm-4v-plus has the worst SR and SPL, but it performs the best in terms of NE (Navigation Efficiency). By analyzing the search process, we find that the GPT-4-o guided searcher can successfully identify the target object when it is at a certain distance, while glm-4v-plus requires the agent to move closer to the target object to recognize it successfully, which reduces the NE.

\begin{table}[!htp]  
    \centering  
    \caption{Ablation study of MLLMs for PRPSearcher.}  
    \begin{tabular}{lcccc}
\hline
\textbf{Method} & \multicolumn{4}{c}{Total}     \\ \cline{2-5} 
                        & SR$\uparrow$ & MSS$\downarrow$& SPL$\uparrow$ & NE$\downarrow$\\ \hline
Qwen-vl-max             &  51.68&  36.07&  40.09&  63.62
\\
glm-4v-plus             &  48.32&  38.04&  39.56&  \textbf{61.68}
\\
GPT4-o                  & \textbf{53.50}& \textbf{35.20}&  \textbf{40.57}&  64.86
\\ \hline
\end{tabular}
    \label{tab:performancellm}  
\end{table}

\subsection{Case Study}
As shown in Fig. \ref{fig:case}, we present a successful case and a failed case of PRPSearcher. 
In the successful case, the MLLM-based agent reasons on the target information to identify related objects, which are then used to build the semantic map and cognitive map for the search. Initially, the agent looks around the surroundings based on exploration advice. Subsequently, it identifies the presence of trees and signs in the scene, assigning them attraction values (0.95 and 0.9). Guided by the 3D cognitive map's exploitation advice, the agent searches a row of shops with trees under a building. Thanks to the denoising mechanism, the agent is able to search along this row of shops and eventually finds the target. \textit{In this case, the denoising mechanism ensures the agent remained focused, ignoring similar shops and successfully finding the target. Crucially, correlating trees with the target in the scene enhances efficiency by guiding the search toward the correct area.}

In a representative failure case, the target is "A black car parked next to the wall." Due to sparse visual information in the target image, the reasoning on this image yields only a few semantic cues: "Car, Windows, Building, Wall." Consequently, PRPSearcher initially prompts the UAV agent toward buildings within the environment. After verifying that encountered vehicles are incorrect, the UAV agent follows exploration advice to explore the space, and subsequently discovers additional buildings. But ultimately, the search terminates unsuccessfully since the search exceeds the step limit. Notably, among all baseline methods evaluated, only human agents and the FBE method locates the target. \textit{This case underscores limitations in PRPSearcher's spatial exploration efficiency and highlights the gap in its spatial semantic reasoning relative to human abilities.}
\section{Conclusion}
In this study, we introduced a relatively unexplored Autonomous Visual Object Search (AVOS) task for UAVs in complex urban environments. We formalized the AVOS task and introduced CityAVOS, the first dedicated benchmark dataset featuring diverse urban objects and scenarios, facilitating standardized evaluation. To tackle this task, we proposed a novel agentic method, namely PRPSearcher, which pioneers a three-tier cognitive architecture mimicking human perception, reasoning, and planning through specialized semantic, cognitive, and uncertainty maps. Also, we introduced an IPT prompting mechanism to guide the UAV agent to balance exploration and exploitation during the action planning. The experimental results demonstrate PRPSearcher's significant advantages over existing methods in both search efficiency and success rate. This work represents a substantial step towards enabling embodied UAV target search capabilities in complex city spaces. 
In the future, we will attempt to further improve PRPSearcher by incorporating collaborative human-agent or multi-agent strategies to handle more complex AVOS tasks (e.g. long-horizon multi-target search).


\bibliographystyle{ACM-Reference-Format}
\bibliography{sample-base}

\clearpage
\appendix

\section{Appendix}

In this appendix, we present detailed information on the dataset, methodology, and experimental procedures as well as results to enhance readers' understanding of our work.

\subsection{Details on Dataset}
\label{app.d1}

The collection process of the CityAVOS dataset can be described as follows.

\begin{figure}[!htp]
  \centering
   \includegraphics[width=0.9\linewidth]{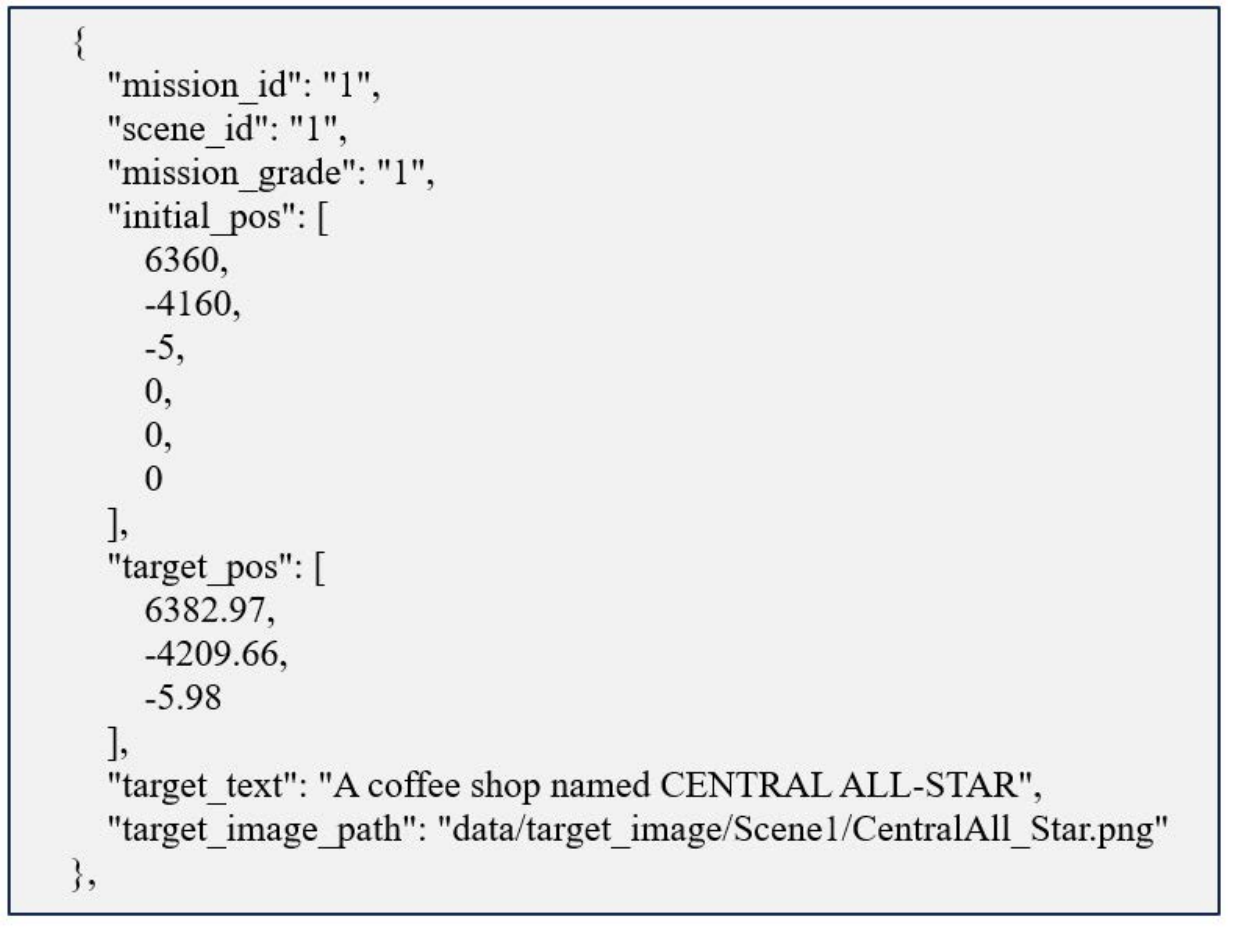}
   \caption{The task in the CityAVOS dataset.}
   \label{fig:t1}
\end{figure}

\begin{itemize}[topsep=0pt,parsep=0pt]
    \item  \textbf{Environment Modification}:
We modified the urban environment in EmbodiedCity by introducing target objects specifically designed for AVOS tasks.

    \item \textbf{Scene Delimitation}:
Define the boundaries of the scene and determine the step size based on the overall scene range and the dimensions of the target objects. Set the starting point for the search task within the scene.

    \item \textbf{Task Generation}:
Identify and locate the target objects. Capture images of each target and its surrounding context. Prepare corresponding target descriptions and classify them based on difficulty. The task example is shown in Fig. \ref{fig:t1}.

    \item \textbf{Trajectory Collection:}:
Develop Python scripts to control the drone and enable automated path collection for trajectory acquisition.

    \item \textbf{Manual Verification}:
Each trajectory is manually reviewed to identify and filter out incorrect paths. Any erroneous trajectories are then regenerated manually.

\end{itemize}

\begin{table}[!htp]  
    \centering  
    \caption{Task Classification and basis based on task difficulty.}
    \renewcommand\arraystretch{1.2}
    \resizebox{\linewidth}{!}
    {
    \begin{tabular}{cccc}
\hline
\textbf{Task Difficulty}   & Easy                                                                                  & Medium                                                                                                            & Hard                                                                                                                        \\ \hline
\textbf{Task Attributes}   & \begin{tabular}[c]{@{}c@{}}Easy to search \\ and easy to identify\end{tabular}        & \begin{tabular}[c]{@{}c@{}}Hard to search \\ and easy to identify\end{tabular}                                    & \begin{tabular}[c]{@{}c@{}}Hard to search \\ and hard to identify\end{tabular}                                        \\
\midrule
\textbf{Scene Size}        & Small                                                                                 & Large                                                                                                             & Large                                                                                                                       \\
\textbf{Goal Uniqueness }  & Unique                                                                                & Unique                                                                                                            & Non-Unique                                                                                                                  \\
\textbf{Number of Tasks}   & 1320                                                                                  & 720                                                                                                               & 380                                                                                                                         \\
\midrule
\textbf{Examples of Tasks} & \begin{tabular}[c]{@{}c@{}}Search for the\\  cafe shop \\ on this street\end{tabular} & \begin{tabular}[c]{@{}c@{}}Search for the Industrial and\\  Commercial Bank of China\\ near the park\end{tabular} & \begin{tabular}[c]{@{}c@{}}Search for the garbage\\  station next to the parking \\ space in this neighborhood\end{tabular} \\ \hline
\end{tabular}
    }
    \label{tab:datasetdif}  
\end{table} 

Tab. \ref{tab:datasetdif} shows the classification rules and task examples. We aim to comprehensively evaluate the agent's ability to identify targets, explore spatially, and perform cognitive reasoning in the AVOS tasks through these three different difficulty levels. Specifically, in the easy tasks, the object is unique in a small scene, requiring the agent to possess basic semantic understanding and spatial exploration abilities. In the medium tasks, the object is unique in a large scene, which demands that the agent explore the space efficiently. In the hard tasks, the object is non-unique in a large scene, necessitating the agent to perform comprehensive reasoning and decision-making based on the characteristics of the object and its surrounding environment.

\subsection{Details on PRPSearcher Approach}
\label{app.approach}

\begin{figure}[!htp]
  \centering
   \includegraphics[width=0.9\linewidth]{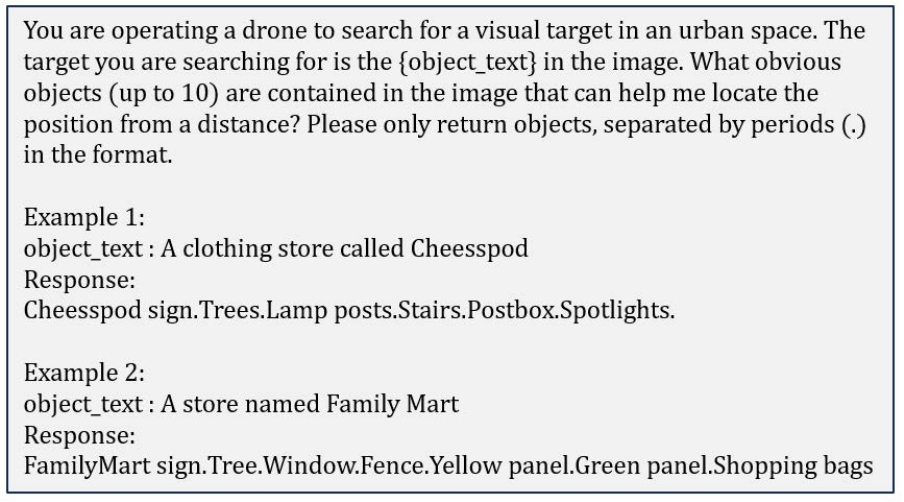}
   \caption{Prompt for related semantics.}
   \label{fig:s1}
\end{figure}

\textbf{Details of object-centric semantic segmentation.}  
Preparing a segment prompt for the semantic segmentation model helps control the semantic scope during the segmentation process. Object-centric semantic segmentation first leverages an MLLM to infer semantics related to the target object, then feeds the related semantics into the segmentation model. This approach effectively reduces the computational complexity during subsequent semantic map construction. The prompt $Prompt_{rel}$ input to an MLLM for this process is shown in Fig. \ref{fig:s1}.

\begin{figure}[!htp]
  \centering
   \includegraphics[width=0.9\linewidth]{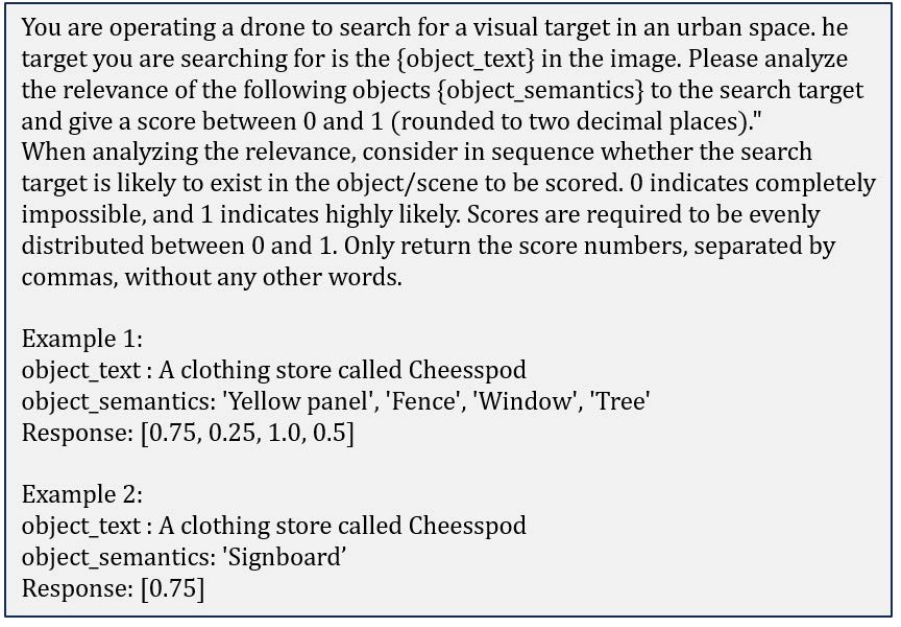}
   \caption{Prompt for 3D cognitive maps.}
   \label{fig:s2}
\end{figure}

\textbf{Details on attractions in 3D cognitive map.}  
The 3D cognitive map reflects the attraction of scene semantics to the agent, which essentially stems from the relevance between the semantics and the target object. To obtain the semantics and their corresponding attractions, the agent needs to perform reasoning using an MLLM. The prompt used for this reasoning process is shown in Fig. \ref{fig:s2}.

\begin{figure}[!htp]
  \centering
   \includegraphics[width=0.9\linewidth]{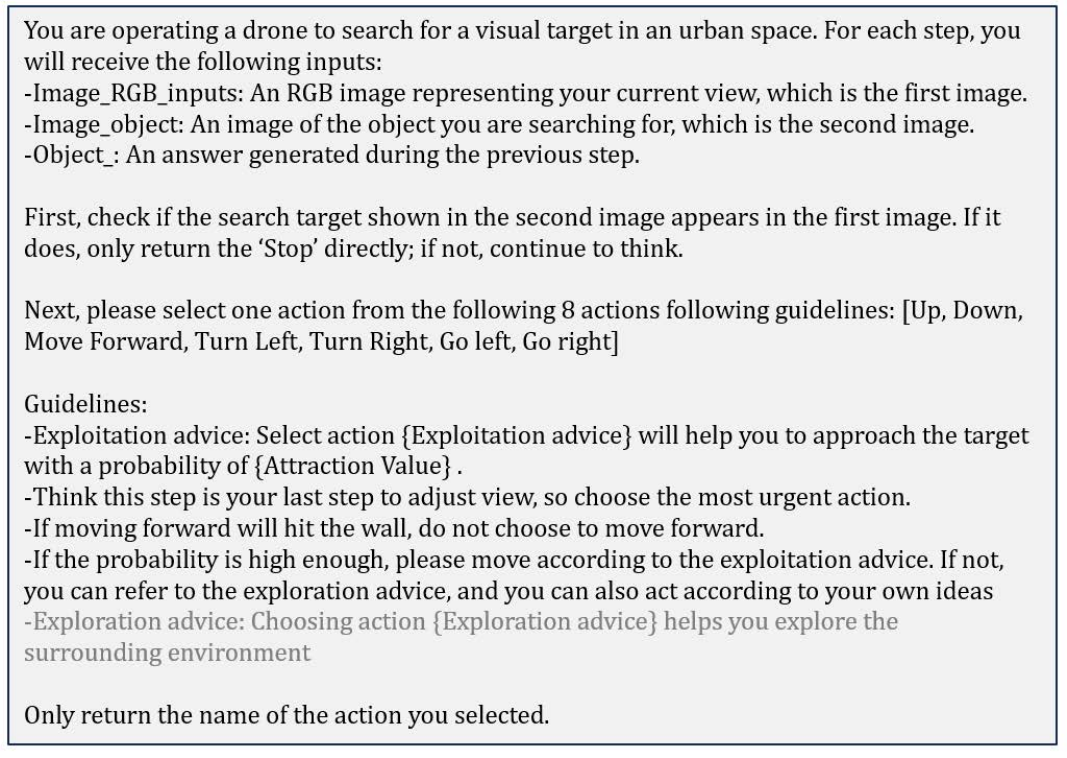}
   \caption{Prompt for action planning.}
   \label{fig:s3}
\end{figure}

\textbf{Details of action planning.} 
In our approach, the agent's action planning is guided by an MLLM, with the corresponding prompt illustrated in Fig. \ref{fig:s3}. The attraction score is used as a probabilistic cue for adopting exploitation advice, while the frequency of exploration advice in the prompt is modulated by the parameter $\theta$ to realize the IPT mechanism. During the action planning process, the MLLM-based agent also needs to determine whether the target object has been found based on the RGB image from the current viewpoint and execute the stop action accordingly.

\begin{figure*}[!htp]
  \centering
   \includegraphics[width=\linewidth]{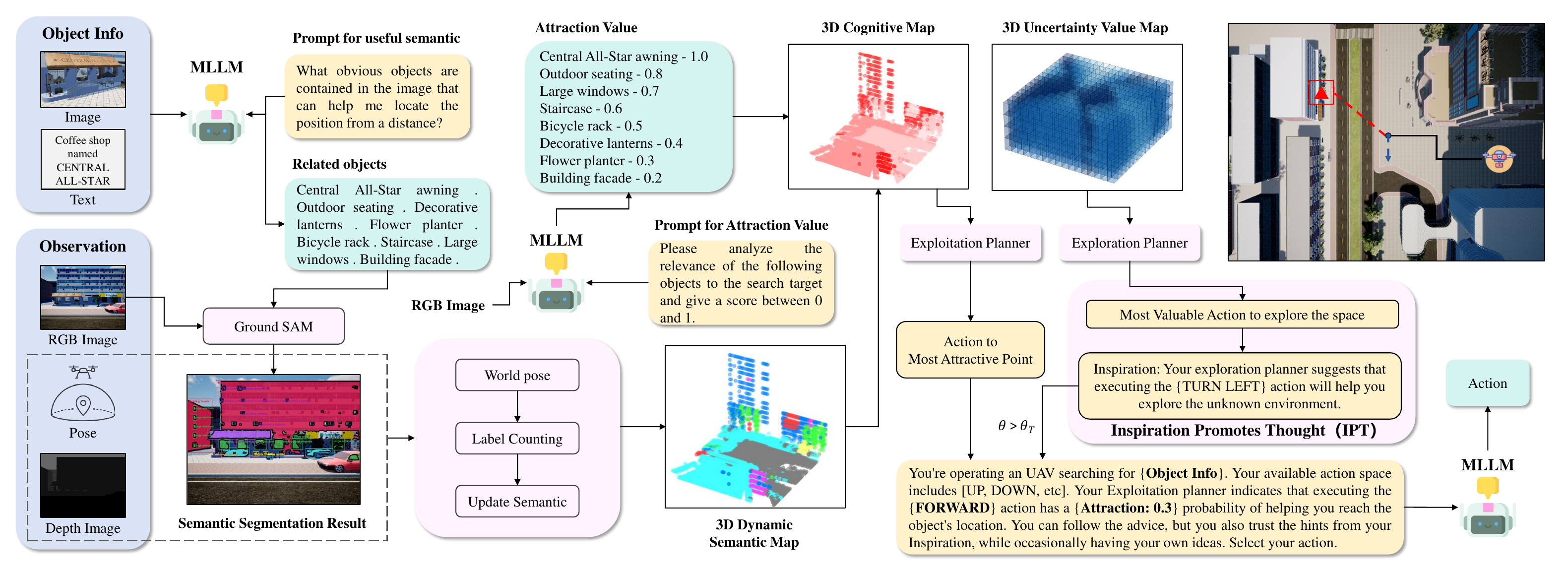}
   \caption{Workflow of the proposed approach—PRPSearcher.}
   \label{fig:workflow}
\end{figure*}

\textbf{Details of approach workflow.}
To better understand our approach, we illustrate the workflow in Fig. \ref{fig:workflow}. Before commencing the object search, the MLLM-based agent performs reasoning based on the given object information to identify objects related to the target.

During the search process, the drone continuously captures RGB images and depth maps from its current pose at each step. The agent first updates a 3D dynamic semantic map using these visual inputs. This involves performing semantic segmentation on the RGB images, where pre-identified related objects serve as prompts for Grounded SAM to produce object-centric semantic segmentation results. The resulting masks and labels are then fused with the depth data to compute world coordinates, which are used to dynamically update the semantic map.

The agent then constructs a 3D cognitive map through further reasoning. It evaluates the correlation between observed semantic elements and the target object, assigning an attraction value to each object, which quantifies how strongly an object attracts the agent's attention within the scene. By mapping these attraction values to their respective semantic elements, the agent forms the 3D cognitive map. Simultaneously, the drone updates a 3D uncertainty map, reducing the uncertainty values of regions within its current field of view.

Finally, both exploitation advice (from the cognitive map) and exploration advice (from the uncertainty map) are generated. These outputs are integrated through the IPT prompt mechanism to effectively guide the agent’s action planning.

\subsection{Details on Experiments}
\label{app.experiment}
\subsubsection{Metrics}
\label{app.metric}
The formulations of the four metrics are presented as follows. Consider a set $ER =\{er_1, er_2, ...er_q\}$ that contains the results of $q$ experiments, where each element $er_i$ is a four-tuple $er_i=\{fs_i, ss_i, tl_i, fp_i\}$. Here, $fs_i$ is a Boolean flag, with $fs_i=1$ indicating that the UAV successfully located the target object in the i-th experiment, and $fs_i=0$ otherwise. The variable $ss_i$ denotes the number of search steps taken, $tl_i$ represents the length of the search trajectory, and $fp_i$ indicates the final position of the UAV when the search ceased in the i-th experiment. For this set of experimental results $ER$, SR and MSS can be calculated using the following formula:
\begin{equation}   
SR={\sum\nolimits_{i=1}^{q}{f{{s}_{i}}}}/{q}
\end{equation}  
\begin{equation}   
MSS={\sum\nolimits_{i=1}^{q}{s{{s}_{i}}}}/{q}
\end{equation}  

Given the ground-truth of the target position $tp^*$ and the length search trajectory $tl^*$, SPL and NE can be calculated as:

\begin{equation}   
NE=\sum\nolimits_{i=1}^{q}{\left\| f{{p}_{i}}-fp_{i}^{*} \right\|}/q
\end{equation}  
\begin{equation}   
SPL=SR \centerdot \sum\nolimits_{i=1}^{q}{t{{l}_{i}}/tl_{i}^{*}}
\end{equation}  

\subsubsection{Baselines}

\begin{itemize}[leftmargin=*]
    \item \textbf{Random Exploration (RE):} At each step of the search process, the UAV randomly selects a feasible action from the action space. An action is deemed feasible if it keeps the UAV within the scene boundaries and avoids collisions with any obstacles. The UAV continues to use visual input to detect the presence of the target object and executes the "Stop" action upon successful identification.
    \item \textbf{Frontier-Based Exploration (FBE):} The UAV continues moving forward until it nears the boundary of the environment or encounters an obstacle. It then performs a turning maneuver to proceed with the search along the perimeter. Given the three-dimensional nature of the environment, random vertical movements are introduced to enhance the search process.
    \item \textbf{L3MVN:} We used GPT-4o as the LLM and VLM in this algorithm. Since the original algorithm was designed for indoor environments in a two-dimensional space, we modified the corresponding 2D components when adapting the code for a three-dimensional urban environment. Specifically, we replaced the 2D semantic map with a 3D semantic map and integrated it with the frontier map. Additionally, we adjusted the global policy to better suit the CityAVOS task.
    \item \textbf{WMNav:} We used Gemini 1.5 Pro as the VLM in this algorithm. Since the UAV in our environment can only obtain first-person view images, we modified the WMNav algorithm accordingly to ensure fairness in the comparative experiments. Specifically, the algorithm was adapted to make predictions based on first-person visual input and to construct the curiosity value map from this perspective.
    \item \textbf{STMR:} We used GPT-4o as the LLM and VLM in this algorithm. Since the code for this algorithm has not been open-sourced, we reproduced the algorithm based on our understanding of the technical approach described in the paper.
    \item \textbf{Human Agent:} At each step of the algorithm’s execution, we presented the human participants with an image of the target object along with its corresponding textual description. Based on the first-person view from the drone, participants were asked to select an action from a predefined set of possible actions. When a participant believes the target has been located, they select the "Stop" action to terminate the current task.
\end{itemize}

\subsubsection{Experiment Configuration}

Our code is executed in a Python 3.9 environment. The experiments are conducted on a Windows 10 platform equipped with an Intel i7-14700KF CPU and an NVIDIA GeForce RTX 4070 Ti SUPER GPU.

\subsubsection{Large Model Configuration} 

All the MLLMs used in this experiment were accessed via API calls. The API endpoints are as follows: GPT-4o (https://openai.com/index/hello-gpt-4o/), Qwen-VL-Max (https://dashscope-intl.aliyuncs.com), and GLM-4V-Plus (https://open.bigmodel.cn/api/paas/v4/chat/completions).

\subsubsection{Case Study}
\begin{figure}[htp]
  \centering
   \includegraphics[width=0.8\linewidth]{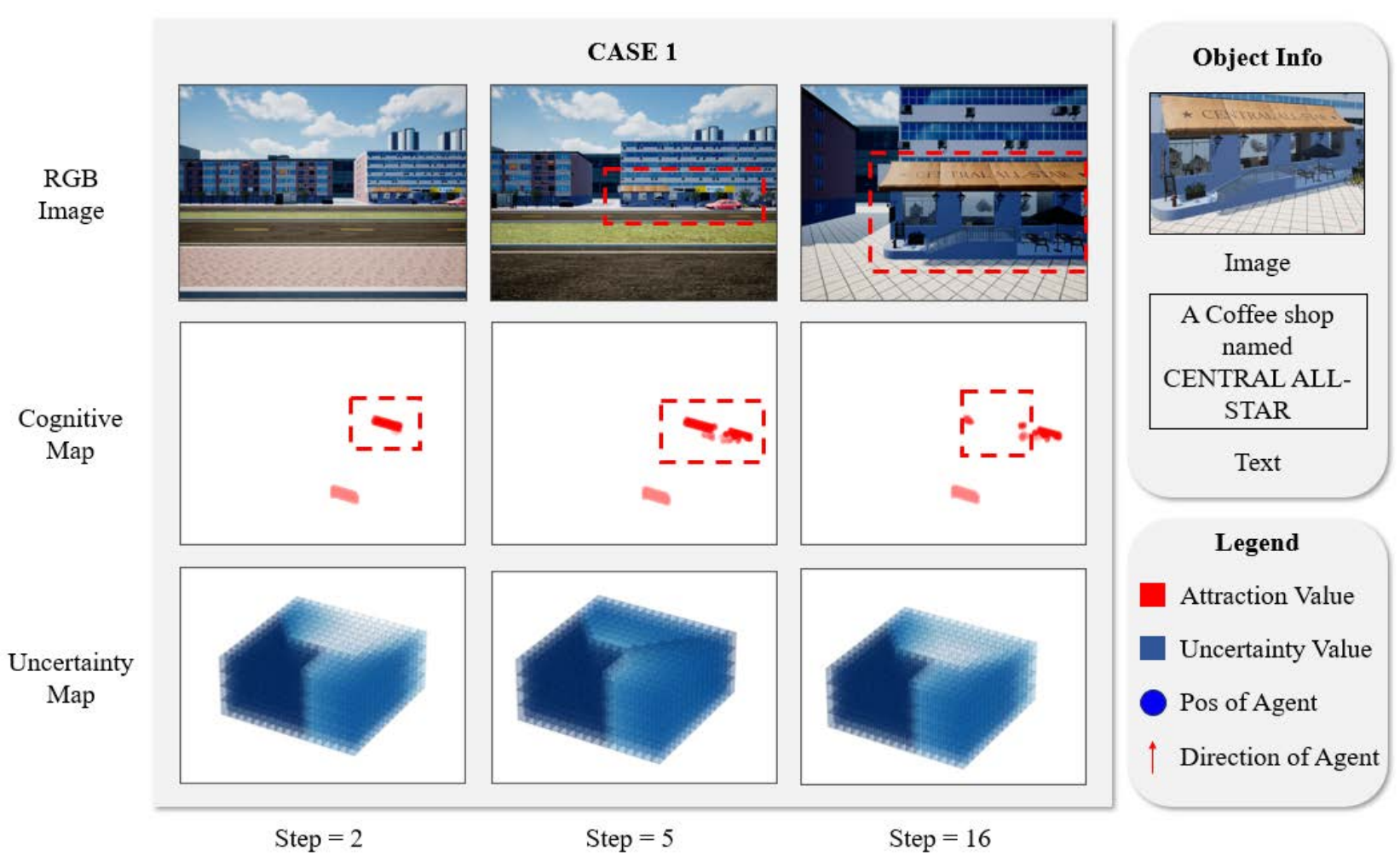}
   \caption{Running process of typical case 1.}
   \label{fig:case1}
\end{figure}

\begin{figure}[htp]
  \centering
   \includegraphics[width=0.8\linewidth]{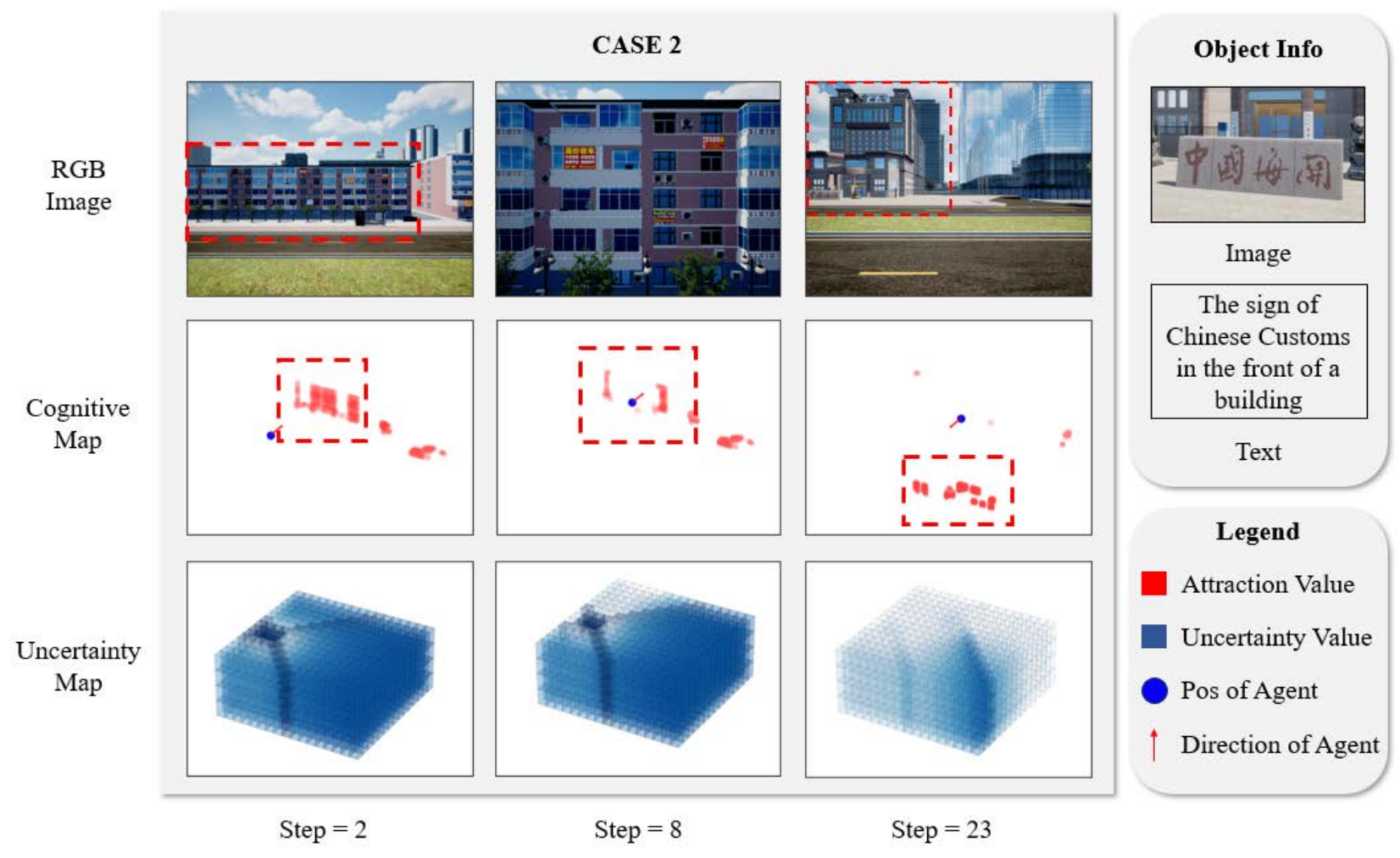}
   \caption{Running process of typical case 2.}
   \label{fig:case2}
\end{figure}

Below we show the illustrative runs of selected episodes. In Fig. \ref{fig:case1} and \ref{fig:case2}, we can observe the mapping process of the cognition map and the uncertainty map based on observations.

Case 1: In this scenario, the task assigned to the UAV agent is to search for a coffee shop named CENTRAL ALL-STAR within an urban environment, which is a relatively straightforward search case. In the second step, the agent identifies a row of shops situated at the base of a building and consequently assigns high attraction values to this region within its 3D cognitive map. Guided by exploitation advice, the agent proceeds towards this area. Upon close approach, it successfully recognizes the target object and executes the "Stop" action. Subsequent verification confirms the correctness of the detection, and the search task is considered successful.

Case 2: In contrast, the second case involves a more complex search task, wherein the UAV agent is required to locate the signage of the Chinese Customs office positioned in front of a building. As depicted in the figure, the agent initially detects multiple signs at the base of a nearby building. However, due to the limited resolution at a distance, it is unable to immediately determine their relevance to the target. The agent thus navigates toward these high-attraction areas to perform a closer inspection. Leveraging the denoising mechanism, the agent is capable of effectively filtering out irrelevant objects. In the subsequent search steps, the agent adopts the exploration advice, investigating previously unvisited regions. Ultimately, the agent successfully identifies the target signage. Although this task requires more search steps compared to Case 1, the target is nonetheless located successfully, demonstrating the robustness of the proposed method.

\end{document}